\documentclass{article}
\pdfoutput=1
\usepackage{spconf,amsmath,graphicx,hyperref}
\usepackage{subfig}
\usepackage{comment}
\usepackage[dvipsnames]{xcolor}
\title{LLMs as Layout Designers: Enhanced Spatial Reasoning for Content-Aware Layout Generation}
\name{%
  Sha Li$^{1}$,
  Stefano Petrangeli$^{2}$,
  Yu Shen$^{2}$,
  Xiang Chen$^{2}$,
  Naren Ramakrishnan$^{1}$,
}
\address{%
  $^{1}$Virginia Tech \\
  $^{2}$Adobe Research
}
\usepackage{amsmath}
\usepackage{svg}
\usepackage{caption}
\usepackage{array} 
\usepackage{amssymb}
\usepackage{pifont}
\usepackage{enumitem}
\setlist[itemize]{itemsep=0pt, parsep=0pt, topsep=0pt, partopsep=0pt}
\usepackage{booktabs}
\usepackage{multirow}
\usepackage{graphicx} % 

\usepackage{xspace}

\def\lays{\textit{LaySPA}\@\xspace}

\def\eg{\textit{eg}\@\xspace}
\begin{document}
%\ninept
%
\maketitle
\begin{abstract}

While Large Language Models (LLMs) have demonstrated impressive reasoning and planning abilities in textual domains and can effectively follow instructions for complex tasks, their ability to understand and manipulate spatial relationships remains limited. Such capabilities are crucial for content-aware graphic layout design, where the goal is to arrange heterogeneous elements onto a canvas so that final design remains visually balanced and structurally feasible. This problem requires precise coordination of placement, alignment, and structural organization of multiple elements within a constrained visual space. To address this limitation, we introduce \lays, a reinforcement learning–based framework that augments LLM-based agents with explicit spatial reasoning capabilities for layout design. \lays employs hybrid reward signals that jointly capture geometric constraints, structural fidelity, and visual quality, enabling agents to navigate the canvas, model inter-element relationships, and optimize spatial arrangements. Through group-relative policy optimization, the agent generates content-aware layouts that reflect salient regions, respect spatial constraints, and produces an interpretable reasoning trace explaining placement decisions and a structured layout specification. Experimental results show that \lays substantially improves the generation of structurally valid and visually appealing layouts, outperforming larger general-purpose LLMs and achieving performance comparable to state-of-the-art specialized layout models.

\end{abstract}
\section{Introduction}
\label{sec:intro}
\textit{Content-aware graphic layout} generation is a fundamental task in creative computation. It aims to automatically arrange visual elements on a given canvas such that the layout is structurally coherent, visually appealing and functionally effective. High-quality layouts are critical in a wide range of applications such as digital advertising, web interface, where they directly impact information clarity, aesthetic perception, and user engagement.

Considerable research has been devoted to generating controllable layouts that balance aesthetic appeal and functional effectiveness. Traditional approaches primarily formulate layout generation as a numerical optimization problem. Generative models, including GANs \cite{li2019layoutgan, zhou2022composition}, VAEs \cite{jiang2022coarse, cao2022geometry}, autoregressive models \cite{horita2024retrieval}, and diffusion-based models \cite{chai2023layoutdm, zhang2023layoutdiffusion, hui2023unifying, li2023relation}, have been widely adopted. These methods predict the positions and sizes of elements to produce visually balanced layouts. While effective in constrained settings, they suffer from several limitations. First, they rely heavily on large-scale annotated datasets, which are costly and labor-intensive to curate. Second, they struggle to capture semantic and spatial relationships among elements,  resulting in misalignment, detrimental overlaps, or inconsistent hierarchies. Moreover, their outputs are constrained by the distribution of training data, limiting adaptability to diverse or novel design contexts. Retrieval-based enhancements \cite{lin2023layoutprompter, horita2024retrieval, forouzandehmehr2025cal} partially mitigate data scarcity by grounding generation on exemplars from external layout databases, yet they remain largely image-centric, prone to local optima, and offer limited interpretability of design decisions.
\begin{figure}[h!]
  \centering
  \subfloat[Input canvas\label{fig:canvas}]{
    \includegraphics[width=0.15\textwidth]{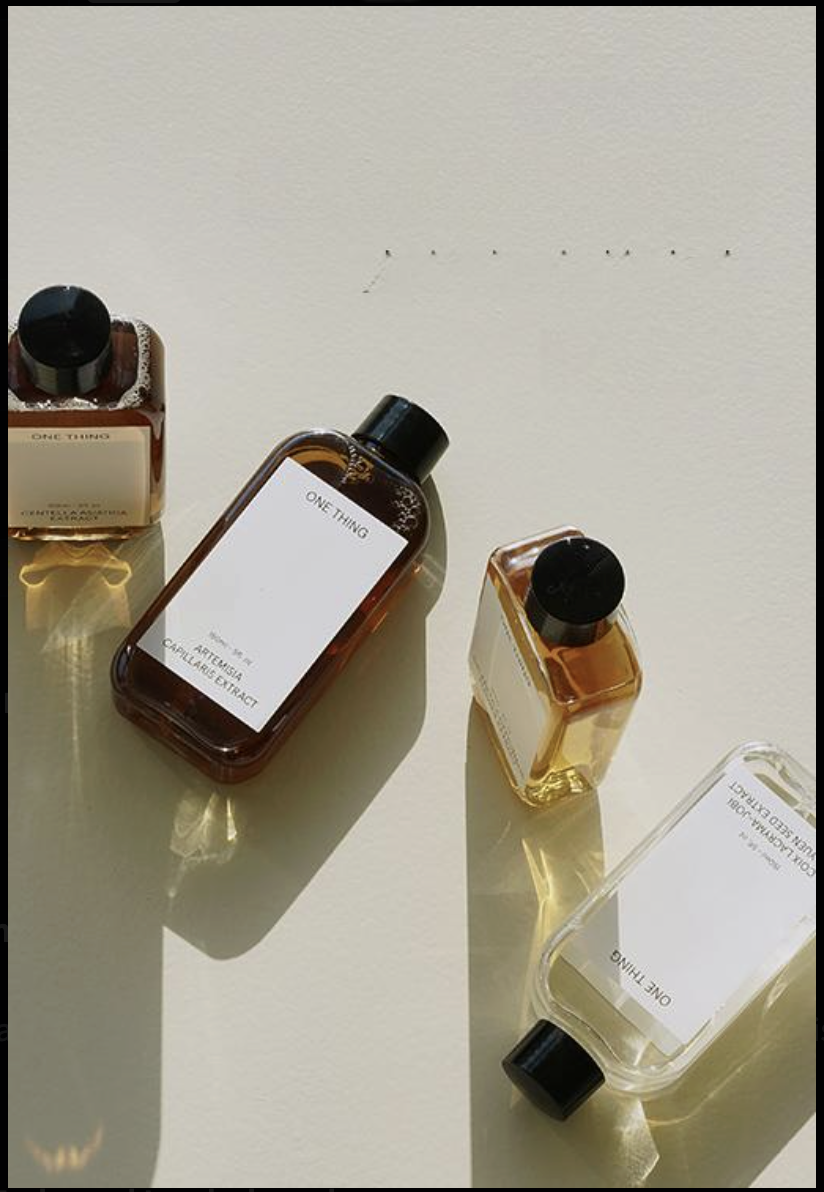}
  }
% optional spacing between figures
  \subfloat[GPT-5 generation \label{fig:ex_gpt5}]{
    \includegraphics[width=0.15\textwidth]{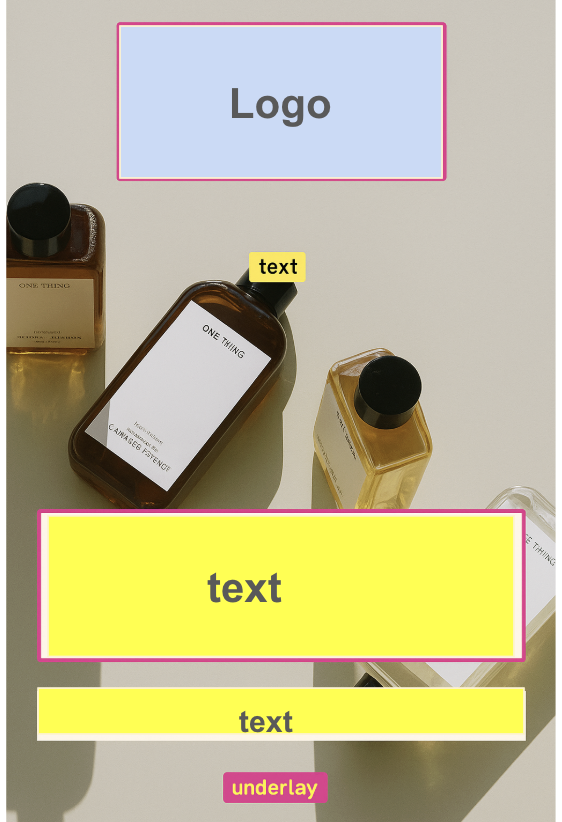}
  }
    \subfloat[Ground truth poster\label{fig:ex_gt}]{
    \includegraphics[width=0.15\textwidth]{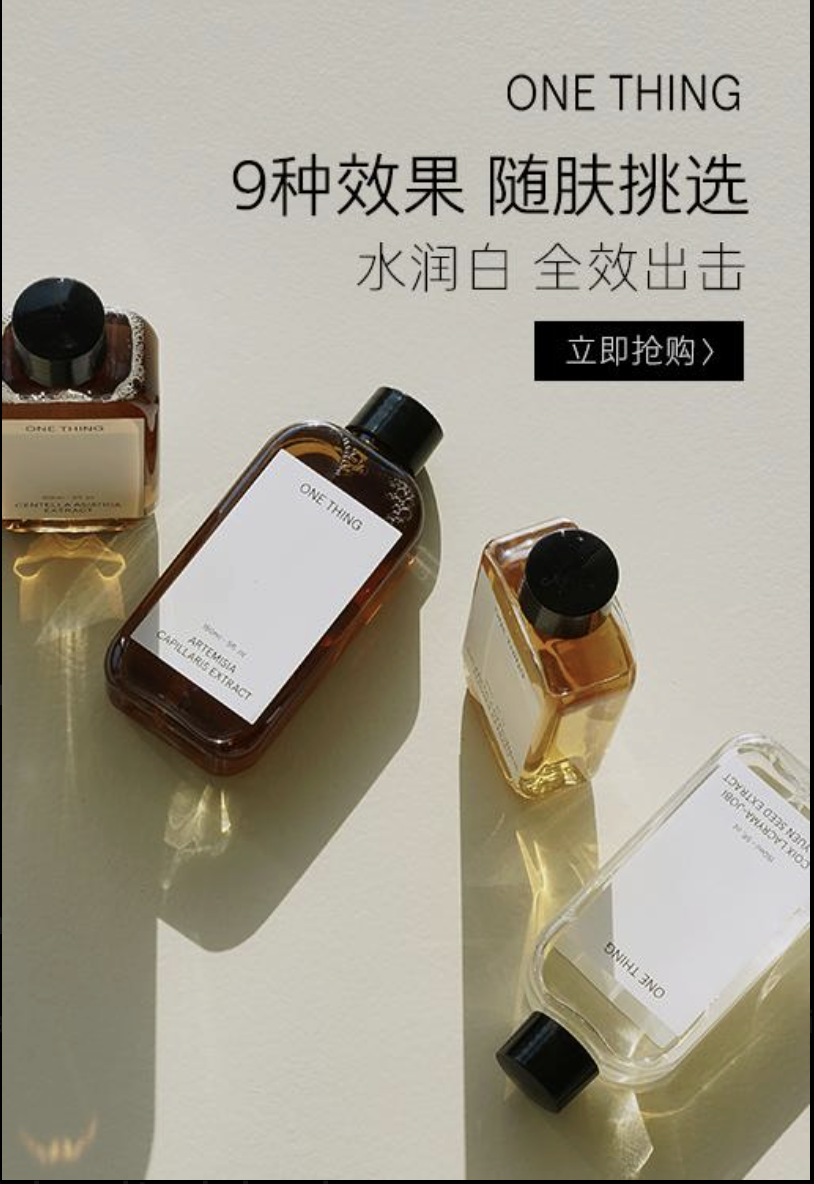}
  }

  \caption{Given the input canvas and a set of elements (one logo, three text, and one underlay), GPT-5 is prompted to generate a poster where elements avoid saliency areas and the underlay decorates one text box. As shown, GPT-5 fails to correctly place elements to avoid salient regions, maintain structural coherence, and position the underlay for decoration.}
  \label{fig:intro_example}
\end{figure}

The emergence of Large Language Models (LLMs) introduces new opportunities for layout generation by leveraging their implicit design knowledge and reasoning capabilities \cite{tang2023layoutnuwa, feng2023layoutgpt}. Recent studies have explored translating layouts into text-based representations to exploit LLMs’ flexibility in sequence modeling. However, these approaches primarily depend on intrinsic capabilities such as in-context learning (ICL), chain-of-thought (CoT) reasoning, and prompt engineering, without explicitly modeling spatial relationships or structural constraints at either the element or canvas level. Consequently, even state-of-the-art LLMs often struggle to produce layouts that preserve structural consistency, geometric plausibility, and visual coherence. 
Figure~\ref{fig:intro_example} illustrates this limitation. Given an input canvas and a predefined set of elements—one logo, three text blocks, and one underlay, we instructed GPT-5 to generate a poster that avoids salient image regions and positions the underlay beneath one text block for decorative emphasis. The resulting design reveals clear spatial deficiencies: GPT-5 fails to consistently avoid salient regions, struggles to maintain element alignment and balanced distribution, produces inconsistent element sizes, and misplaces the underlay relative to the intended text. These shortcomings are striking when compared with the human-designed ground-truth poster. This observation highlights two fundamental challenges for LLM-based layout design. \textit{(C1) Spatial cognition deficiency.} Layout design requires nuanced spatial reasoning to align elements, maintain global structural coherence, and respect geometric constraints. LLMs lack intrinsic spatial understanding, making it difficult for them to capture multi-object alignment and hierarchical relationships, which are critical for high-quality layouts. \textit{(C2) Open-ended design space with limited supervision}. Layouts are inherently diverse, allowing multiple valid and visually appealing configurations for the same set of elements. At the same time, paired canvas-to-layout examples are scarce, limiting the possibilities for explicit mapping-based training and complicating evaluation, as no single ``gold standard” exists.

To address these challenges, we reframe content-aware layout generation as a \textit{policy learning} problem, where an LLM-based agent learns design policies to make design decisions under spatial and structural constraints. We introduce \lays, a reinforcement learning (RL) framework that equips LLM agents with explicit spatial reasoning and adaptive decision-making capabilities to handle diverse design contexts. Our approach integrates two key components that targeted at the challenges outlined above: \textit{(K1) Hybrid reward design (addresses C1)}. We define reward functions that jointly optimize structural feasibility and visual quality, encouraging the agent to capture geometric relationships, align elements, and maintain global layout coherence. \textit{(K2) Self-exploration and dynamic decision-making (addresses C2)}. Layout generation is reformulated as an iterative process, where the agent self-explores and interacts with a spatial evaluation environment, receives stepwise feedback, and incrementally adjusts design policies beyond static memorization or heuristic-based retrieval.

The main contributions of this work are as follows:
\begin{itemize}
    \item We propose a reinforcement learning framework that empowers LLMs with explicit spatial reasoning for content-aware layout generation.
    \item We design dual-layer reward functions integrating geometric and aesthetic principles, guiding the agent to produce layouts that are both structurally valid and visually harmonious.
    \item We demonstrate through experiments that \lays effectively equips LLMs with spatial reasoning capabilities, outperforming existing generative baselines and larger-scale LLMs in producing structurally coherent and visually appealing layouts.
\end{itemize}
\begin{figure*}[htbp!]
\begin{center}
\captionsetup{skip=-5pt} 
  \includegraphics[scale=0.2]{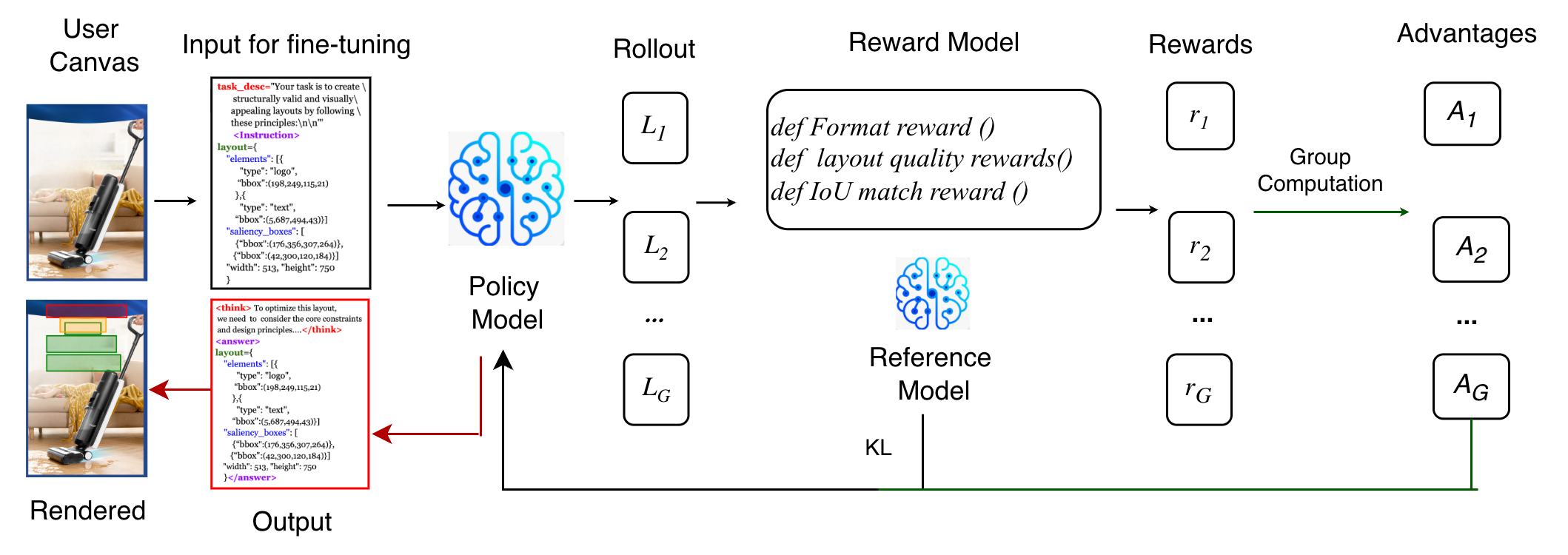}
  \end{center}
  \vspace{-2em}
  \caption{Overview of the \lays framework.}
    \label{fig:overview}
\end{figure*}

\section{Method}
\label{sec:met}

\subsection{Problem Formulation}

The core problem we address is content-aware layout generation: given a background canvas and a set of content elements, how can an LLM learn to model spatial constraints, multi-object relationships, and overall layout structure in order to generate designs that specify the number, type, and positions of elements (\eg., text, logos, underlays) while maintaining both structural feasibility and aesthetic quality.

Formally, a layout $\mathcal{L}$ consists of $N$ elements ${e_{i}}_{i=1}^{N}$, where each element $e_{i}=(x_{i}, y_{i}, w_{i}, h_{i}, c_{i})$ is defined by its top-left coordinates $(x_{i}, y_{i})$, width and height $(w_{i}, h_{i})$ and category $c_{i}$. The canvas may also include a set of saliency regions $S = \sum_{j=1}^{M}{s_j}$, with each $s_j=(x_j, y_j, w_j, h_j, c_j)$ denoting visually important areas derived from saliency detection. The task is to infer appropriate positions and sizes $(x_i, y_i, w_i, h_i)$ for each element given its category $c_i$ and the canvas's saliency context $S$.

A fundamental challenge in layout generation is the inherent multiplicity of valid configurations: for a given set of elements, there exist numerous structurally plausible and aesthetically appealing layouts. This diversity makes traditional supervised fine-tuning (SFT) inadequate, as learning from a single ground-truth layout cannot capture the open-ended design space. To address this, we formulate layout generation as a policy learning problem. An LLM agent learns a design policy $\pi_\theta$ that iteratively decides element placements under structural and spatial constraints, receiving evaluative signals from a feedback environment rather than relying solely on explicit canvas-layout mappings. 

\subsection{Training Framework }

\lays adopts a reinforcement learning paradigm in which an LLM-based agent learns layout policies through trial-and-error interactions with a spatial evaluation environment. Figure~\ref{fig:overview} presents an overview of the framework. Given a canvas containing a background image and a set of elements with predefined types, \lays first identifies salient regions and encodes both the canvas and its elements into a compact JSON-based representation. The agent then generates multiple rollouts, each corresponding to a candidate layout that includes both an interpretable reasoning trace and a structured JSON specification. These candidates are assessed by a hybrid reward model across three complementary dimensions:
\textit{(1) Format correctness}, ensuring that outputs conform to the required reasoning–layout schema;
\textit{(2) Structural constraints}, evaluating geometric validity such as boundary adherence, non-overlap, and size consistency; and
\textit{(3) Visual quality}, measuring alignment, spacing, hierarchy, and overall element distribution. The feedback is used to optimize the agent’s policies via \textit{Group Relative Policy Optimization} (GRPO)~\cite{shao2024deepseekmath}, which facilitates adaptive and generalizable spatial reasoning without relying on rigid supervision. Ultimately, \lays produces two complementary outputs: (i) an interpretable reasoning trace that exposes the agent’s spatial reasoning and decision-making process, and (ii) a structured layout representation consisting of bounding boxes with associated element categories.

\subsection{Hybrid Reward Design}

\label{sec:funcs}
To steer the agent toward generating structurally valid and visually appealing layouts, we design a hybrid reward scheme that integrates three complementary objectives: output validity, structural plausibility, and visual quality. All reward components are normalized to $[0,1]$, where higher values indicate better performance. 
%The final reward is comprised by:
%$R=\lambda_{v}R_{v}+\lambda_{s}R_{s}+\lambda_{q}R_{q}$, where %$\lambda$ represnets the weights for different reward components. 

\textbf{Format Reward} ($R_{format}$) \ 
This component evaluates whether the model's output follows the expected reasoning-layout structure. Each response must include a $<think>$ block for the agent’s reasoning steps and an $<answer>$ block containing a JSON-formatted layout. Rewards are assigned hierarchically: 0.1 if either block is missing, 0.2 if the JSON is unparsable, 0.5 if the JSON is valid but element types or counts mismatch the input, and 1.0 if both blocks are present, the JSON is valid, and all element types and counts match the specification. This graded scheme prevents the model from hallucinating or omitting elements while ensuring that the output can be reliably assessed by downstream reward components.

\textbf{Layout Quality Reward} ($R_{\tiny{quality}}$) \   
This component evaluates layouts based on geometric constraints and visual design principles, combining five sub-metrics normalized to [0,1], where higher scores indicate better performance. 

\begin{figure}[htbp]
\centering
\renewcommand{\arraystretch}{1.2} % adjust vertical spacing
\setlength{\tabcolsep}{3pt}      % adjust column spacing
\begin{tabular}{m{0.2\linewidth}|m{0.35\linewidth}|m{0.35\linewidth}}

\centering Collision $R_{\tiny{icr}}$ &
\includegraphics[width=0.99\linewidth]{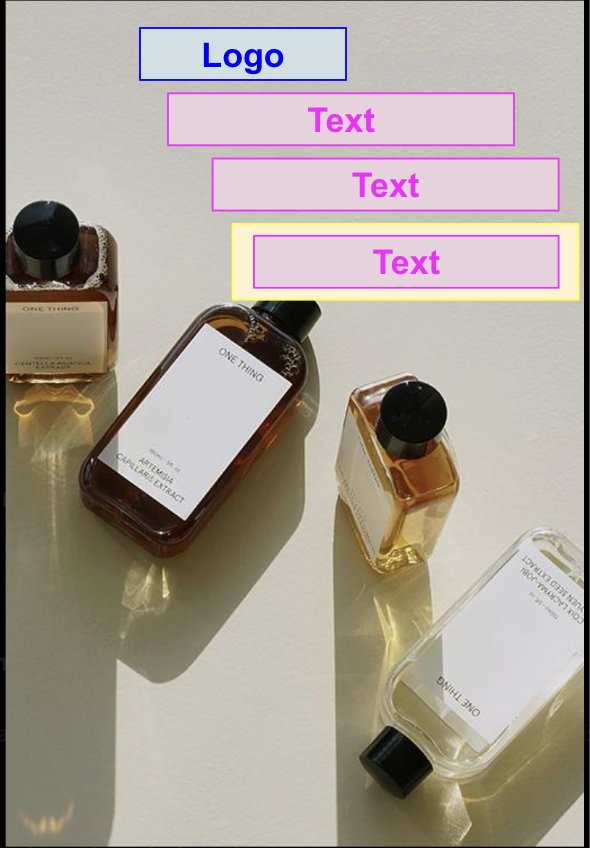} &
\includegraphics[width=0.99\linewidth]{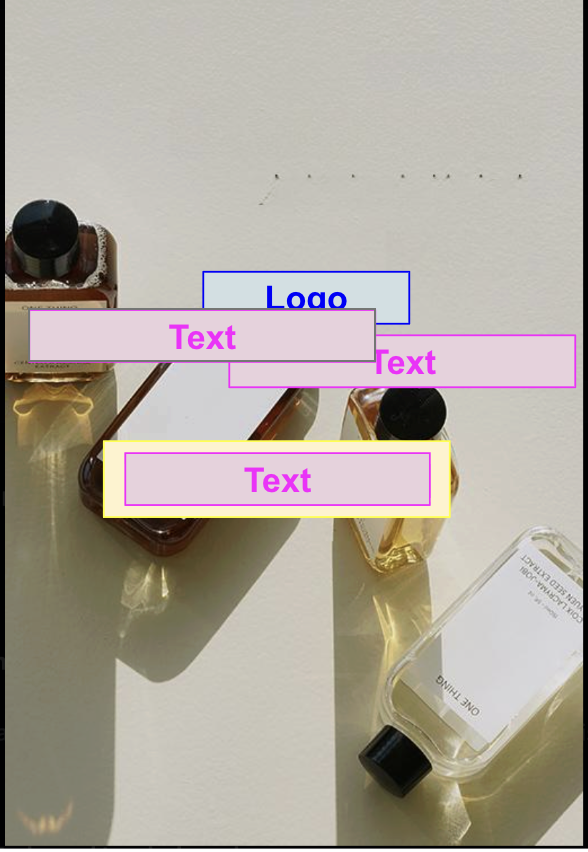} \\
\hline
\centering Alignment $R_{\tiny{al}}$ &
\includegraphics[width=0.99\linewidth]{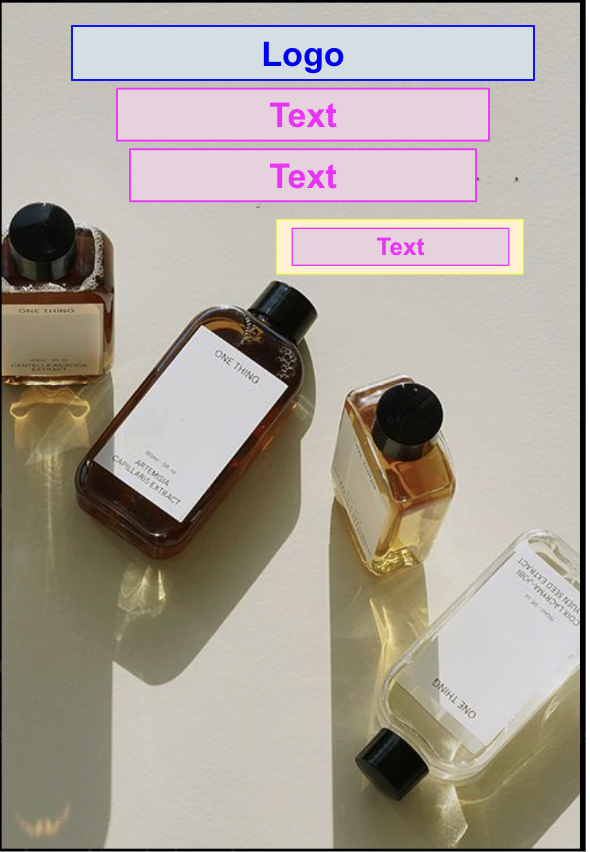} &
\includegraphics[width=0.99\linewidth]{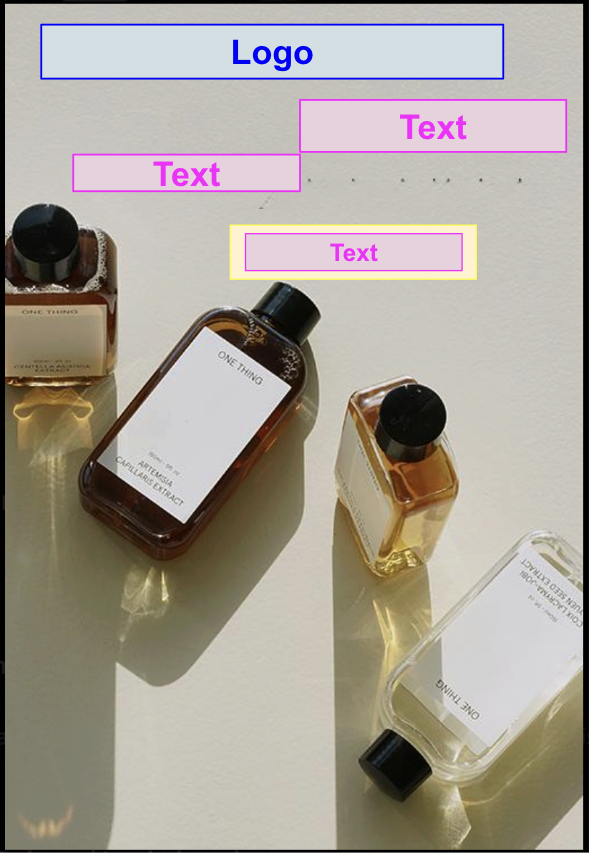} \\
\hline
\centering Distribution $R_{\tiny{dis}}$ &
\includegraphics[width=0.99\linewidth]{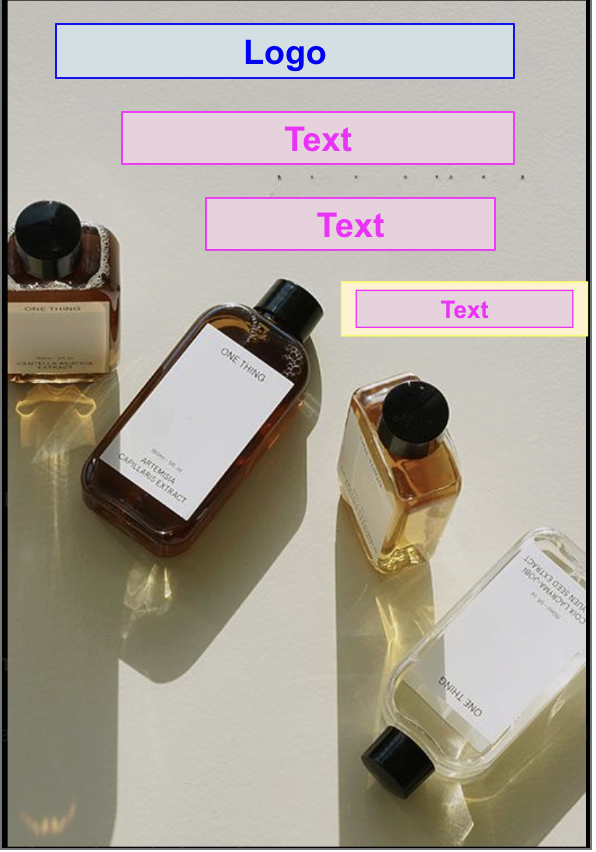} &
\includegraphics[width=0.99\linewidth]{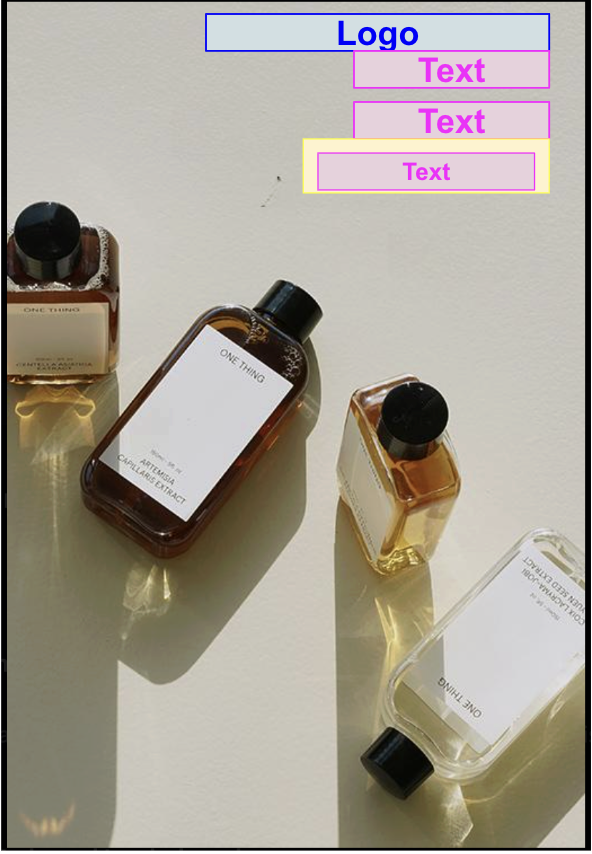} \\
\hline
\centering Spacing $R_{\tiny{sp}}$ &
\includegraphics[width=0.99\linewidth]{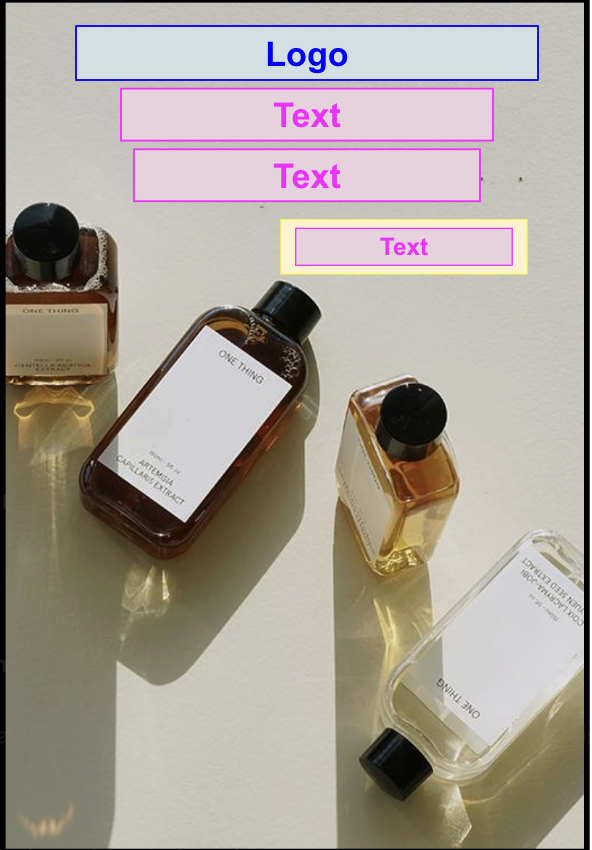} &
\includegraphics[width=0.99\linewidth]{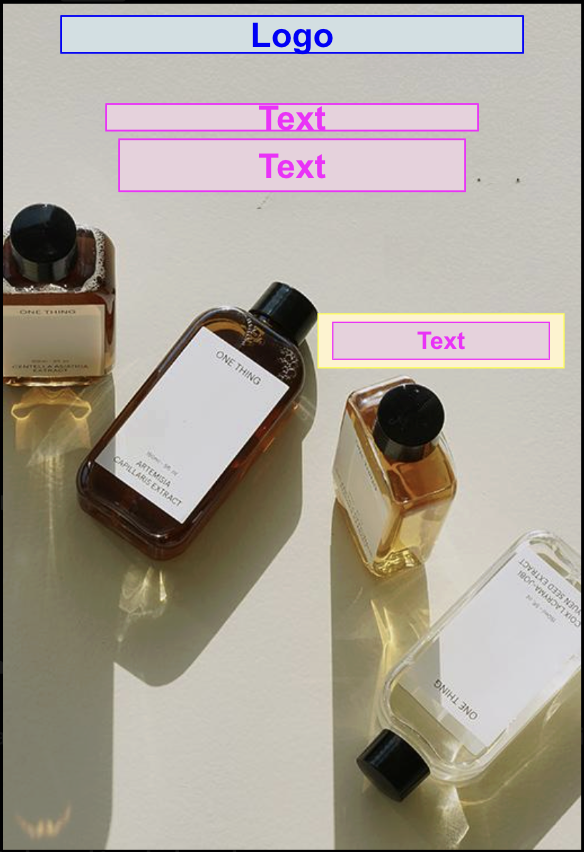} \\
\hline
\centering Underlay-text $R_{\tiny{ut}}$ &
\includegraphics[width=0.99\linewidth]{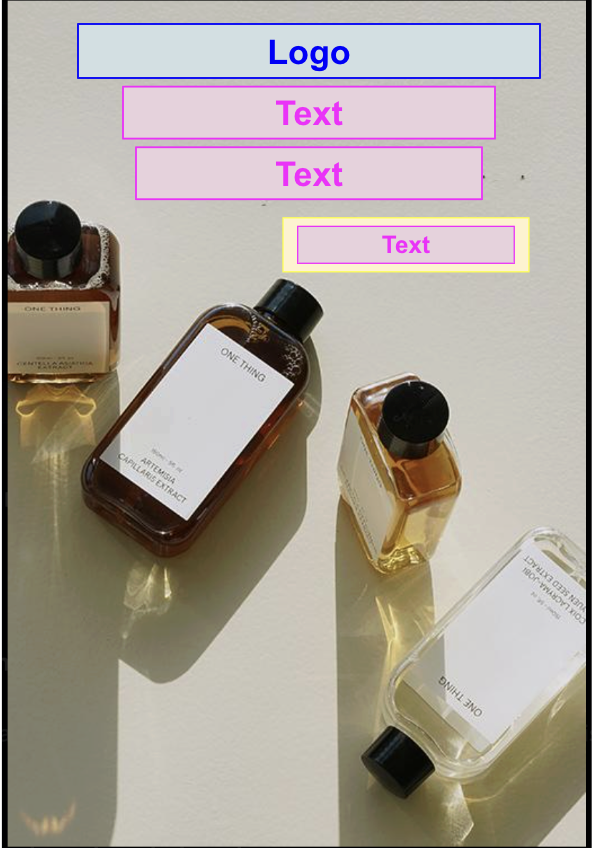} &
\includegraphics[width=0.99\linewidth]{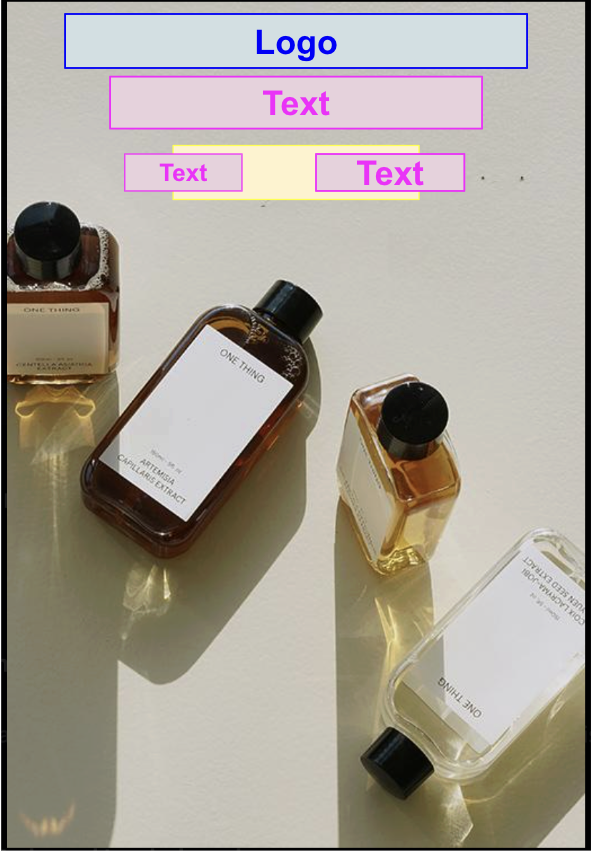} \\
\hline
 &
\centering \textbf{\textcolor{Green}{\ding{52}}} &
\centering \textbf{\textcolor{red}{\ding{56}}}\\

\end{tabular}
\caption{An illustration of how each layout quality score functions guides the agent toward human-preferred designs (\textbf{\textcolor{Green}{\ding{52}}}) while discouraging (\textbf{\textcolor{red}{\ding{56}}}) undesirable spatial arrangements.
}
\label{fig:textimagegrid}
\end{figure}

\textit{Inverse Collision Rate} ($R_{\tiny{icr}}$) \  penalizes detrimental overlaps between non-compatible elements (\eg., text, logos, embellishments and salient regions), while encouraging overlaps between underlays and text (as underlays are intended to decorate text). The collision score is computed by measuring the ratio of the overlapped area to the total area of the involved bounding boxes. The reward is defined as one minus the normalized overlap ratio:$R_{\tiny{icr}}=1- \frac{Area(b_{i}\cap b_{j})}{Area(b_{i})+Area(b_{j})-Area(b_{i}\cap b_{j})}$, where $b$ represents a bounding box and $Area(\centerdot)$ denotes the area of a region. $R_{\tiny{icr}}$ encourages conflict-free placements.

\textit{Alignment Score} ($R_{\tiny{al}}$) measures visual alignment relative to the canvas center and among elements. Element-to-canvas alignment is computed via the distance between the mean element center and the canvas center, while element-to-element alignment uses the variance of element centers. Lower variance yields higher scores.

\textit{Distribution Score} ($R_{\tiny{dis}}$) evaluates how evenly elements are arranged on the canvas based on two measures. The first is a spread variance, computed as the average squared distance of element coordinates from their mean position, higher variance indicates broader dispersion across the canvas. The second is a grid coverage score, obtained by dividing the canvas into a $3*3$ grid and calculating the fraction of cells containing at least one element. The final score is the average of these two measures, balancing spatial spread and grid utilization to encourage efficient coverage while discouraging element clustering.

\textit{Spacing Consistency Score} ($R_{\tiny{sp}}$) quantifies vertical rhythm by computing the normalized variance of distances between adjacent element centers. Elements are first sorted by vertical positions, then the distances between adjacent centers are computed. The variance of these spacings is normalized by the mean spacing and canvas height. Higher scores correspond to more even spacing.

\textit{Underlay-Text Constraint Reward} ($R_{\tiny{ut}}$)  enforces semantic consistency by aligning each underlay with exactly one corresponding text element. Layouts are penalized if multiple text elements overlap a single underlay or if the underlay’s position and size deviate from its associated text. Perfect alignment receives a score of 1, partial matches are assigned intermediate scores, and incorrect pairings are scored 0. This mechanism encourages the agent to respect functional relationships between elements, ensuring both readability and coherent design.

Finally, these sub-rewards are aggregated into a weighted overall layout quality score. Fig.~\ref{fig:textimagegrid} illustrates how these layout quality scores guide the agent in generating content-aware layouts that balance structural feasibility and visual appeal.

\textbf{IoU Matching Score} ($R_{\tiny{IoU}}$) \  
To encourage human-like placement patterns, we incorporate an Intersection-over-Union (IoU) reward that measures the geometric accuracy of generated layouts relative to human-designed ground truth. For each element, IoU is computed as the ratio of the intersection area to the union area between the predicted and reference bounding boxes. Unlike collision or alignment rewards which assess internal structural quality, IoU provides external guidance by aligning model outputs with human design references, which implicitly encode principles of balance, proportion, and functional grouping. Aligning generated bounding boxes with these references guides the agent toward realistic spatial organization and visually coherent structures. 

\textbf{Final Hybrid Score}  \  The final hybrid reward for each generated layout is a weighted combination of the three components:
\begin{equation}
R(L_{i})=\lambda_{f}R_{format} + \lambda_{q}R_{quality} + \lambda_{u}R_{IoU}
\end{equation}
We assign weights $\lambda_{f}=0.1$, $\lambda_{q}=0.8$, and $\lambda_{u}=0.1$ giving priority to layout quality since it spans multiple dimensions, while down-weighting $R_{IoU}$ to promote diversity.

\subsection{Learning Process}

\label{sec:learnprocess}
To optimize the design policy under this hybrid reward, we employ GRPO, which uses group-level comparisons to establish relative baselines, making it more efficient and better suited for exploratory layout generation.

At each training step, given a background canvas and a set of input elements, the policy model samples a group of candidate layouts ${L_{1}, L_{2}, ..., L_{G}}$ from the old policy $\pi_{\theta_{old}}$. Each layout is evaluated by the hybrid reward function described in \S\ref{sec:funcs}, resulting in a set of reward values ${r_{1}, r_{2}, ..., r_{G}}$. The relative quality of each layout within the group is then computed by subtracting the group mean reward, yielding group-based advantages ${A_{1}, A_{2}, ..., A_{G}}$. These advantages reflect whether a layout is better or worse than its peer candidates, ensuring stable gradient signals without requiring an explicit critic. The policy model $\pi_{\theta}$ is then updated by maximizing the GRPO objective:
\begin{equation}
\begin{split}
\mathcal{J}_{\text{GRPO}}(\theta) &= 
\mathbb{E}_{x \sim \mathcal{D}, \{\mathcal{L}_i\}_{i=1}^{G} \sim \pi_{\theta_\text{old}}} \\
&\Bigg[ 
\frac{1}{G} \sum_{i=1}^{G} \frac{1}{|\mathcal{L}_i|} 
\sum_{t=1}^{|\mathcal{L}_i|} 
\min \Bigg\{ 
\frac{\pi_\theta(\mathcal{L}_{i,t} \mid x, \mathcal{L}_{i,<t})}
     {\pi_{\theta_\text{old}}(\mathcal{L}_{i,t} \mid x, \mathcal{L}_{i,<t})} \widehat{A}_{i,t}, \\
&
\text{clip}\Big(
\frac{\pi_\theta(\mathcal{L}_{i,t} \mid x, \mathcal{L}_{i,<t})}
     {\pi_{\theta_\text{old}}(\mathcal{L}_{i,t} \mid x, \mathcal{L}_{i,<t})}, 1-\varepsilon, 1+\varepsilon
\Big) \widehat{A}_{i,t}
\Bigg\} 
\Bigg] \\
&- \beta \, \mathcal{D}_{KL}\big[\pi_\theta \,\|\, \pi_\text{ref}\big]
\end{split}
\end{equation}
where $\varepsilon$ and $\beta$ are hyper-parameters, and $\widehat{A}_{i,t}$ denotes the advantage, computed from the relative rewards of outputs within each group.

\section{Experiments}
\label{sec:exp}
\textbf{Datasets} \ 
We evaluate \lays on two benchmarks for content-aware layout generation: CGL \cite{zhou2022composition} and PKU-PosterLayout \cite{hsu2023posterlayout}. The PKU dataset consists of text, logo, and underlay elements, while CGL includes an additional embellishment category. For training and evaluation, we randomly sample 3k instances from the 48.4k annotated training examples in CGL and 3k instances from the 7.97k examples in PKU. Testing is conducted on 6.06k samples from CGL and 997 samples from PKU. Saliency maps are generated using off-the-shelf methods \cite{qin2019basnet, qin2022highly} and subsequently converted into bounding boxes.

\textbf{Layout Representation}  \  Layouts are represented in JSON format. To enable prediction, element coordinates and sizes are masked with the special token \texttt{[MASK]}, prompting the LLM to replace them with the corresponding predicted positions and dimensions. 

\textbf{Experimental Setup}  \  We adopt Qwen-2.5-7B-instruct \cite{qwen2.5} as backbone models for our framework. Both models are fine-tuned with LoRA \cite{hu2022lora}. 

\textbf{Baselines and Settings}  \ We compare \lays against representative generative and LLM-based approaches:
\begin{itemize}
    \item \textbf{DS-GAN} \cite{hsu2023posterlayout}: a non-autoregressive generative model combining CNN and LSTM architectures to model layout distributions.
    \item \textbf{PosterLlama} \cite{seol2024posterllama}: an LLM-based method that formulates layout generation as HTML code generation to leverage reasoning over structured design elements.
    \item \textbf{Qwen-2.5-7B-Instruct without applying \lays}: backbone LLMs evaluated in a zero-shot setting, used to assess whether performance improvements arise from the \lays framework rather than the underlying model capacity.
    \item \textbf{GPT-4o} \cite{hurst2024gpt}: a large closed-source LLM evaluated zero-shot without task-specific fine-tuning.
\end{itemize}
 
\textbf{Evaluation Metrics} \ 
We assess \lays from two dimensions: (1) improvement over the base LLM without fine-tuning, to quantify the effectiveness of hybrid reward-guided policy learning, and (2) performance relative to state-of-the-art generative and LLM-based layout methods. For internal learning evaluation, we report layout-specific measures reflecting structural quality: collision rate, alignment score, spacing consistency, and element distribution. These metrics assess how effectively \lays learns spatial reasoning via the proposed hybrid reward scheme. For comparison with SOTA methods, we follow prior works \cite{hsu2023posterlayout, horita2024retrieval}, using the following graphic- and content-focused metrics: \textit{Overlay (Ove $\downarrow$)} measures the average geometric overlap between all pairs of elements excluding underlays; \textit{Underlay Effectiveness (Und $\uparrow$)}  evaluates how well underlay elements support non-underlay content, and \textit{Occlusion (Occ $\downarrow$)} measures the extent to which layout elements overlap salient regions of the canvas. 

\begin{comment}
both 
textbf{Graphic metrics}. These measure the quality of a layout based on the spatial and geometric relationships among its elements, independent of the background canvas. \textit{Overlay} (Ove $\downarrow$) measures the average geometric overlap between all pairs of elements, excluding underlays. \textit{Underlay Effectiveness} (Und $\uparrow$) evaluates how well underlay elements support non-underlay content. \textit{Utility} (Uti $\uparrow$) assesses how effectively the layout utilizes regions suitable for element placement.
 
\textbf{Content metrics}. These metrics evaluate how harmoniously the generated layout elements relate to the canvas. \textit{Occlusion} (Occ $\downarrow$)  measures the extent to which layout elements overlap important regions in the saliency map. 

\end{comment}

\section{Results and Analysis}
\label{sec:results}

\subsection{Quantitative Results}
\label{sec:quant_eval}
\textbf{Learning Effectiveness} \  Table~\ref{tab:learn} presents the structural and aesthetic evaluation scores for the CGL and PKU datasets, comparing the base Qwen models with their fine-tuned counterparts using \lays. Across both datasets, fine-tuning with \lays yields consistent and substantial improvements across all metrics. Notably, the larger Qwen-7B model demonstrates more pronounced gains, with format correctness increasing by 14\%, alignment improving by 63\%, spacing consistency rising by 73\%, distribution enhanced by 26\%, and collision rate reduced by 36\%. These results indicate that \lays effectively guides the LLM-agent to internalize spatial reasoning principles over the canvas, producing structured outputs that respect both geometric constraints and visual design principles.

\begin{table}[h]
\centering
\setlength{\tabcolsep}{1pt}
\caption{Learning effectiveness evaluation on (a) CGL and (b) PKU datasets. Metrics include format correctness (Format↑), collision rate (Coll↓), alignment (Align↑), spacing consistency (Spacing↑), and distribution (Distr↑). \lays demonstrates substantial improvements in structural validity and visual layout quality. \textbf{Bold} numbers indicate the best performance, confirming \lays’s effectiveness in enhancing layout quality.
%\vspace{-0.7em}
}
\subfloat[CGL Dataset]{
\begin{tabular}{l@{\hskip 1pt}ccccc}
\toprule
Model & Format↑ & Coll↓ & Align↑ & Spacing↑ & Distr↑ \\
\midrule
%Qwen-3B & 0.418 & 0.727 & 0.209 & 0.192 & 0.180 \\
%Qwen-3B+\lays  & 0.468 & 0.671 & 0.342 & 0.320 & 0.216 \\
Qwen-7B & 0.873 & 0.692 & 0.319 & 0.326 & 0.253 \\
Qwen-7B+\lays & \textbf{0.998} & \textbf{0.431} & \textbf{0.597} & \textbf{0.569} & \textbf{0.317} \\
\bottomrule
\end{tabular}
}

\subfloat[PKU Dataset]{
\begin{tabular}{l@{\hskip 0.5pt}ccccc}
\toprule
Model & Format↑ & Coll↓ & Align↑ & Spacing↑ & Distr↑ \\
\midrule
%Qwen-3B& 0.403 & 0.727 & 0.210 & 0.195 & 0.187 \\
%Qwen-3B+\lays& 0.469 & 0.633 & 0.445 & 0.325 & 0.217 \\
Qwen-7B& 0.873 & 0.689 & 0.479 & 0.353 & 0.261 \\
Qwen-7B+\lays & \textbf{0.999} & \textbf{0.442} & \textbf{0.659} & \textbf{0.607} & \textbf{0.332} \\
\bottomrule
\end{tabular}
}
\label{tab:learn}
\end{table}

\textbf{Compared with SOTA}  \  Table~\ref{tab:sota} reports results on the CGL and PKU datasets. Real Data corresponds to the annotated samples from the original datasets. Among all compared methods, PosterLlama achieves the strongest overall performance, with an average of 0.0028 on overlap, 0.9958 on underlay effectiveness and 0.1782 on occlusion. This superior performance can be attributed to its integration of visual encoders, its architecture explicitly optimized for layout generation, and the incorporation of strong task-specific priors. Qwen-7B fine-tuned with \lays ranks second, showing substantial gains over its base model, particularly in reducing overlaps and improving underlay utilization. GPT-4o, despite its larger scale, delivers moderate results, highlighting its limited ability to capture fine-grained spatial relations. These results demonstrate that while \lays enables LLMs to acquire effective spatial reasoning capabilities and close the gap with specialized generative models, dedicated architectures like PosterLLaMA, which has visual encoders to capture spatial features of canvas, remain advantageous for achieving the highest performance in layout generation.

Notably, while PosterLLaMA is trained on the full datasets (48,438 samples for CGL and 7,979 for PKU), \lays attains comparable performance using only 3,000 annotated samples. This data efficiency highlights the ability of \lays to empower general LLMs for content-aware layout design, narrowing the gap to specialized generative models. 

\begin{table*}[h]
\centering
\caption{Comparison of overlap (Ove↓), underlay effectiveness (Und↑), and occlusion (Occ↓) metrics on the CGL and PKU datasets, with best results highlighted in bold and second-best results underlined. $+$ denotes base model with \lays. }
\setlength{\tabcolsep}{5pt}  
\begin{tabular}{cccc|cccc}
\toprule
 & \#Params & \multicolumn{3}{c}{(a) CGL Dataset} & \multicolumn{3}{c}{(b) PKU Dataset} \\
\cmidrule(lr){3-5} \cmidrule(lr){6-8}
Model & & Ove ↓ & Und ↑ & Occ ↓ & Ove ↓ & Und ↑ & Occ ↓ \\
\midrule
\textcolor{gray}{Real Data} & - & \textcolor{gray}{0.0003} & \textcolor{gray}{0.9926}& \textcolor{gray}{0.1379} & \textcolor{gray}{0.0013} & \textcolor{gray}{0.9974} & \textcolor{gray}{0.1828} \\
DS-GAN  & 30M & 0.0361 &0.6309 & 0.1521 & 0.0336 & 0.7613 &0.2574 \\
\small PosterLLama  & 7B & \textbf{0.0024} & \textbf{0.9918} & \textbf{0.1476} & \textbf{0.0032} &\textbf{0.9998} & \underline{0.2087} \\
GPT-4o  & 200B & 0.0365 & 0.5873 & 0.1591 & 0.0371 & 0.6384 & 0.2743 \\
%Qwen-3 & 3B & 0.0618 & 0.4993 & 0.2105 & 0.0624 & 0.492 & 0.2601 \\
%Qwen-3B+   & 3B & 0.0481 & 0.5753 & 0.1899 & 0.0385 & 0.5901 & 0.2398 \\
Qwen-7 & 7B & 0.0474 & 0.5729 & 0.1615 & 0.0479 & 0.6059 & 0.2384 \\
Qwen-7B+ & 7B & \underline {0.0257} & \underline{0.6989} & \underline{0.1487} & \underline{0.0260} & \underline{0.7688} & \textbf{0.2072} \\
\bottomrule
\end{tabular}
\label{tab:sota}
\end{table*}

\subsection{Qualitative Results}
\label{sec:qual}
\begin{figure*}[htbp!]
  \centering
  \subfloat[CGL\label{fig:cgl}]{
    \includegraphics[width=0.495\textwidth]{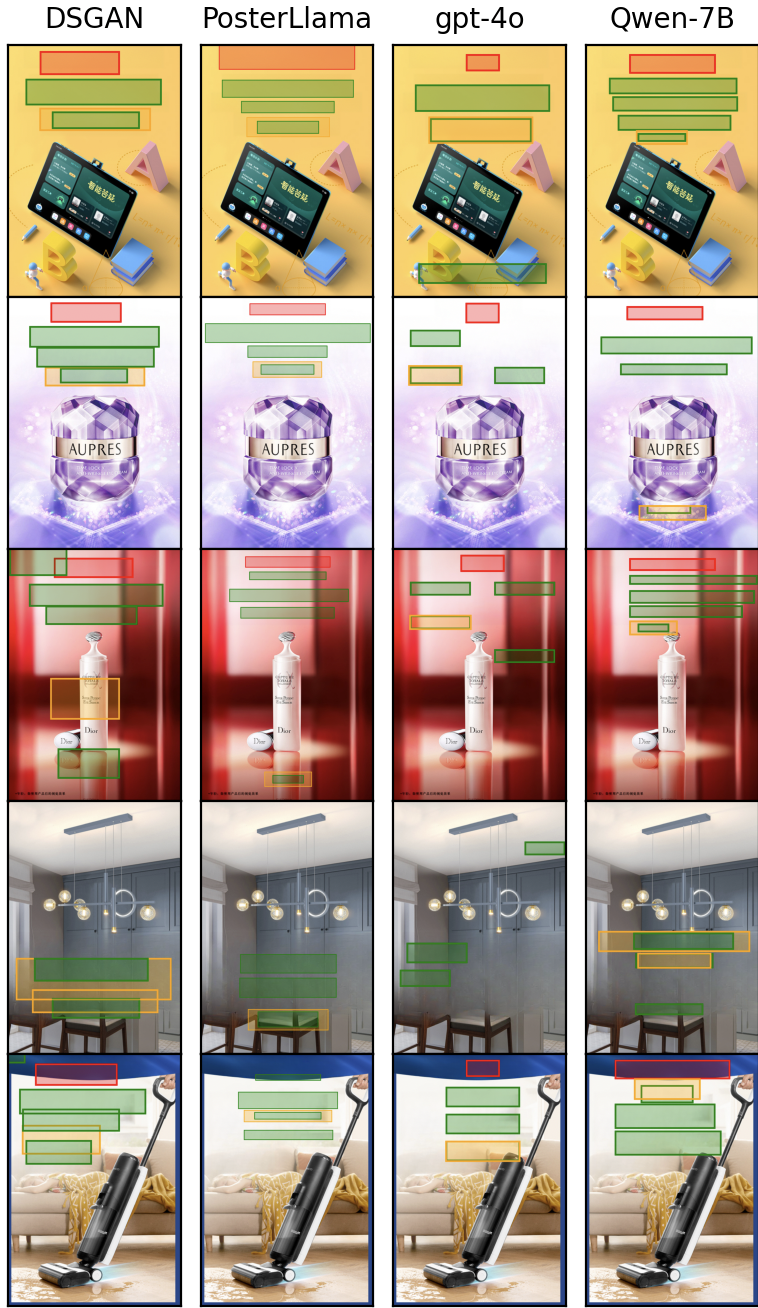}
  }
% optional spacing between figures
  \subfloat[PKU\label{fig:pku}]{
    \includegraphics[width=0.498\textwidth]{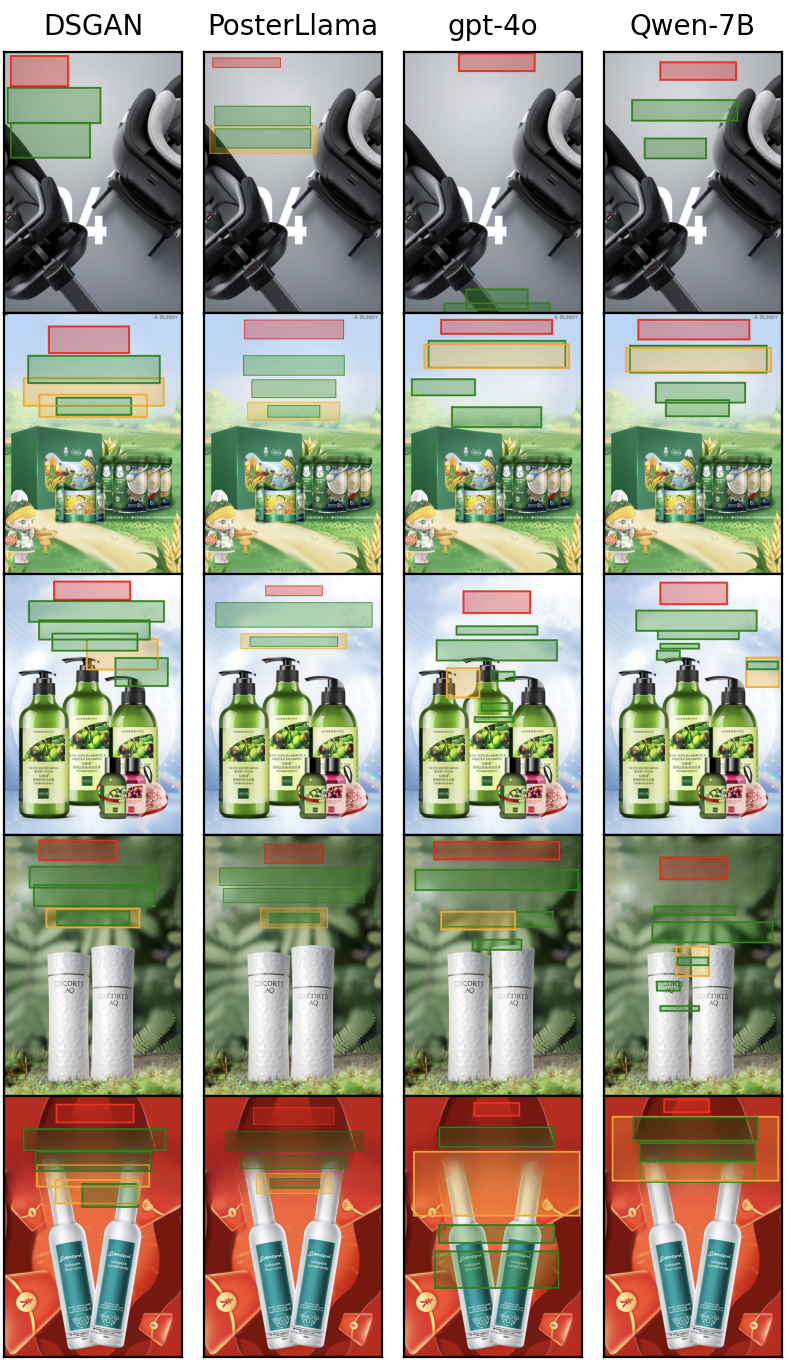}
  }

  \caption{Visualization of generated layouts using different methods on (a) CGL and (b) PKU datasets. Qwen-7B results are generated with \lays fine-tuning. }
  \label{fig:combined}
\end{figure*}

We present qualitative comparisons on the CGL and PKU datasets across different layout generation methods in Figure~\ref{fig:combined}. DS-GAN demonstrates effective canvas utilization and produces elements with relatively consistent sizes; however, its layouts often suffer from geometric flaws such as misalignment and collisions, together with occasional underlay or text placements that interfere with salient background regions. PosterLlama achieves the most visually balanced results overall, delivering layouts that better preserve structural coherence and maintain harmony across element categories. Despite the large sizes of parameters, GPT-4o performs poorly, particularly on the CGL dataset, where frequent misalignment and weak spatial organization undermine layout quality. In contrast, Qwen-7B (with \lays) produces more stable and balanced arrangements, demonstrating stronger spatial reasoning capabilities and overall design quality.

\section{Related Work}
\label{sec:related_works}

\subsection{Content-aware Layout Generation}

Deep generative models have long dominated content-aware layout generation, with most approaches falling into three categories: GAN-based, autoregressive, and diffusion-based methods \cite{chai2023two, chai2023layoutdm}. For example, CGL-GAN \cite{zhou2022composition} employs a Transformer-based encoder–decoder to generate layouts from constraints, while DS-GAN \cite{hsu2023posterlayout} combines CNNs and LSTMs to condition on background images and learn distributions of design sequences for poster layouts. Diffusion-based methods, such as LayoutDM \cite{chai2023layoutdm}, leverage a discrete diffusion process with Transformer-based denoising to capture complex spatial dependencies among elements, yielding more diverse and coherent outputs than GAN-based models. To alleviate the challenge of limited training data, retrieval-augmented methods like RALF \cite{horita2024retrieval} enrich layout generation by fetching similar examples from external databases, though they inevitably inherit the biases and limitations of the retrieved layouts. Beyond direct generation, optimization-based refiners such as LayoutRectifier \cite{shen2025layoutrectifier} focus on post-processing, introducing grid alignment strategies and differentiable containment functions to mitigate misalignment, overlap, and occlusion. Despite advances, these approaches formulate layout generation as a numerical optimization problem, which rely heavily on large-scale annotated data, risk losing semantic relationships among elements, and struggle to generalize beyond training distributions. Differing from these methods, our work frames layout design as a reinforcement learning problem, enabling exploration of diverse design possibilities while maintaining structural feasibility.

\subsection{LLM-based Layout Generation}
Recent advances in LLMs have prompted efforts to leverage their implicit design priors and the representational flexibility of language for layout generation. A common strategy is to serialize layouts into text-based formats such as HTML \cite{tang2023layoutnuwa, seol2024posterllama}, CSS \cite{feng2023layoutgpt}, or SVG \cite{hsu2025postero}, thereby enabling LLMs to reason over structured specifications. For example, PosterLlama \cite{seol2024posterllama} formulates poster design as an HTML generation task, translating design intents into structured layouts. PosterO \cite{hsu2025postero} extends this paradigm by introducing hierarchical SVG-like trees, which support more diverse design intents and accommodate non-rectangular elements. Beyond serialization, relation-aware approaches such as ReLayout \cite{tian2025relayout} enrich layouts with explicit annotations (e.g., regions, saliency, margins) and apply recursive decomposition with Relation-CoT reasoning to better capture spatial dependencies and interactions. Retrieval-based methods further enhance performance by grounding generation in exemplar designs: LayoutPrompter \cite{lin2023layoutprompter} selects prompt examples with constraints resembling test cases, while CAL-RAG \cite{forouzandehmehr2025cal} employs an agentic framework that retrieves exemplars, proposes candidate layouts with an LLM, and iteratively refines them using feedback from a vision-language grader. However, existing LLM-based methods either rely heavily on retrieval or implicit priors, or depend on emergent capabilities such as in-context learning \cite{feng2023layoutgpt} and chain-of-thought reasoning. In contrast, our work explicitly activates and enhances LLMs’ limited spatial reasoning capabilities, guiding models to learn structural relations and multi-object spatial dependencies with hybrid geometric and aesthetic rewards. This approach enables the generation of layouts that are both structurally valid and visually harmonious, with interpretable reasoning trajectories, moving beyond reliance on priors or retrieval heuristics.

\begin{comment}
    \begin{figure*}[htbp!]
  \centering
  \subfloat[CGL\label{fig:cgl}]{
    \includegraphics[width=0.492\textwidth]{figs/cgl.png}
  }\hfill
  \subfloat[PKU\label{fig:pku}]{
    \includegraphics[width=0.48\textwidth]{figs/pku.png}
  }
  \vspace{-0.5em}
  \caption{Visualization of generated layouts using different methods on (a) CGL and (b) PKU datasets. Qwen-3B and Qwen-7B results are generated with \lays fine-tuning.}
  \label{fig:combined}
\end{figure*}

\end{comment}

\section{Conclusion}
\label{sec:conclusion}

We presented \lays, a GRPO-based framework that equips LLM agents with explicit spatial reasoning capabilities for content-aware graphic layout generation. Within this framework, the LLM agent self-explores element placement under structural and spatial constraints, guided by a hybrid reward schema that jointly evaluates output validity, structural plausibility, and visual quality. Experimental results demonstrate that \lays effectively learns spatial reasoning, substantially improving layout quality, outperforming larger advanced LLMs, and achieving performance comparable to specialized state-of-the-art models. To the best of our knowledge, this is the first work to investigate re-purposing LLMs as autonomous layout designers from a spatial reasoning perspective. Several promising directions remain for future work. First, \lays currently relies on pre-detected saliency maps, which overlook the rich visual characteristics of the canvas; incorporating visual semantics may yield stronger alignment between design intent and layout structure. Second, extending \lays with multi-turn reinforcement learning refinements could better simulate iterative human design behaviors. Finally, scaling \lays to broader applications such as user interfaces, magazines, and other structured visual media would further demonstrate its generality and practical impact.

% Below is an example of how to insert images. Delete the ``\vspace'' line,
% uncomment the preceding line ``\centerline...'' and replace ``imageX.ps''
% with a suitable PostScript file name.
% -------------------------------------------------------------------------

%\vfill
%\pagebreak

\vfill\pagebreak

\section*{Acknowledgement}
Original images from PKU-PosterLayout dataset by Hsu et al. available at https://huggingface.co/datasets/creative-graphic-design/PKU-PosterLayout, licensed under CC BY-SA 4.0. Original images from CGL dataset by Zhou et al. available at https://huggingface.co/datasets/creative-graphic-design/CGL-Dataset, licensed under CC BY-NC-SA 4.0.

\newpage
\bibliographystyle{IEEEbib}
\bibliography{strings,paper}

\end{document}